\documentclass{article}
\usepackage{spconf,amsmath,graphicx}
\usepackage{array}
\usepackage{amsmath,amssymb}
\usepackage[utf8]{inputenc}

\usepackage{cite}
\usepackage{multirow}
\usepackage{booktabs}
\usepackage{supertabular}
\usepackage{hyperref}
\usepackage{setspace} 
\usepackage{makecell}
\usepackage{amssymb,booktabs,multirow,array,xspace,bm,enumitem, paralist}

\title{Comprehensive facial expression synthesis \\ using human-interpretable language}
\name{\makecell{Joanna Hong, Jung Uk Kim, Sangmin Lee, and Yong Man Ro*}
	\thanks{This work was conducted by Center for Applied Research in Artificial Intelligence (CARAI) grant funded by DAPA and ADD (UD190031RD).}\thanks{*Corresponding author (ymro@kaist.ac.kr).}
	\address{Image and Video Systems Lab, School of Electrical Engineering, KAIST, South Korea}}
\begin{document}
	%
	\maketitle
	\begin{abstract}
		Recent advances in facial expression synthesis have shown promising results using diverse expression representations including facial action units. Facial action units for an elaborate facial expression synthesis need to be intuitively represented for human comprehension, not a numeric categorization of facial action units. To address this issue, we utilize human-friendly approach: use of natural language where language helps human grasp conceptual contexts. In this paper, therefore, we propose a new facial expression synthesis model from language-based facial expression description. Our method can synthesize the facial image with detailed expressions. In addition, effectively embedding language features on facial features, our method can control individual word to handle each part of facial movement. Extensive qualitative and quantitative evaluations were conducted to verify the effectiveness of the natural language.
	\end{abstract}
	\begin{keywords}
		Facial expression image synthesis, natural language, human-interpretable
	\end{keywords}
	\section{Introduction}
	
	Recently, facial expression synthesis has been widely researched on diverse application of computer vision, such as data augmentation, entertainments (\textit{e.g.,} chatbot), and emotions therapy \cite{Therapy}. Rendering photorealistic facial expression images can give significant impact on the field of affective computing which studies on interpreting human affects.
	
	Many face synthesis methods have noticeably shown promising results \cite{StarGAN, ZHOU, GANimation, LAC-GAN}. \cite{StarGAN, ZHOU} showed advances in synthesizing single facial image expressing seven discrete emotions \cite{Ekman}. Despite its promising results, it has limitation on indicating discrete amounts of human expression. This is because the facial expression is more complicated and diverse to be considered as the emotional aspects \cite{MPI}.
	
	\begin{figure}[t]
		\begin{minipage}[b]{1.0\linewidth}
			\centering
			\centerline{\includegraphics[width=7.5cm]{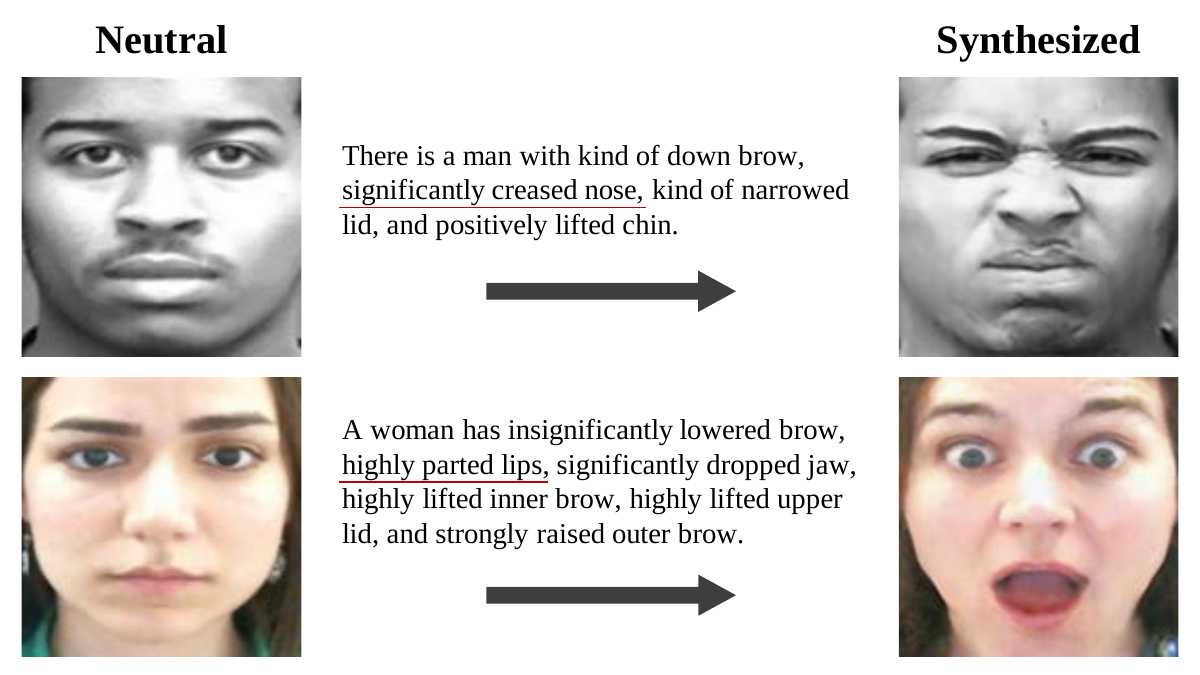}}
		\end{minipage}
		\vspace{-0.75cm}
		\caption{The example synthesized facial expression images from CK+ (top) and DISFA+ (bottom) dataset.}
		\label{fig:1}
		\vspace{-4mm}
	\end{figure}
	
	\cite{GANimation,LAC-GAN} solved these lack of diversities, by proposing models for producing synthetic facial representations using diverse combinations of AUs. These methods utilized facial Action Units (AUs) \cite{FACS}. Since AUs are the fundamental actions numerically categorized from human muscle movements, they can represent more diverse expression with combination of AUs with their intensities. However, there are about 30 major AU categories, so the possible combinations of these categories become extensively large. These extensive amounts of combinations seem hardly related to direct interaction with humans (\textit{i.e.,} ill-intuitive) in applying to the facial expression. Therefore, research on how to easily convey the facial expression representation is essential.
	
	\begin{figure*}[t!]
		\begin{minipage}[b]{1.0\linewidth}
			\centering
			\centerline{\includegraphics[width=13cm]{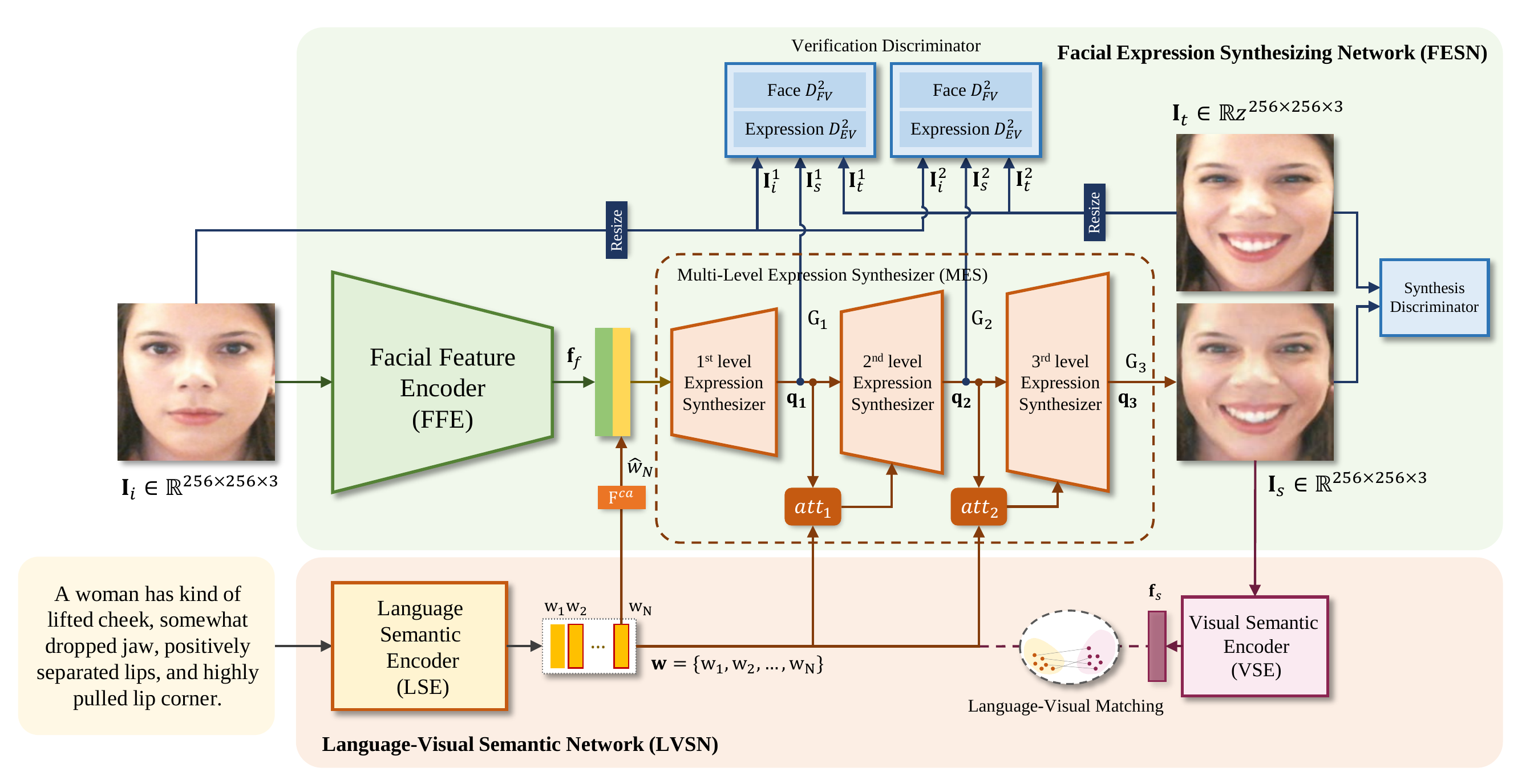}}
		\end{minipage}
		\vspace{-0.95cm}
		\caption{The overall proposed architecture: Facial Expression Synthesizing Network (FESN) (top) and Language-Visual Semantic Network (LVSN) (bottom).}
		\label{fig:2}
		\vspace{-0.595cm}
	\end{figure*}
	
	To alleviate the aforementioned issue, we utilize human-friendly approach for face expression synthesis: use of natural language. Vision and language are already well known to be crucial factors of human intelligence to understand a real world \cite{RogerShank}. Using natural language enables humans to directly control descriptive language expressing details of facial movement. For example, as shown in Fig. 1, such underlined descriptive words, \textit{significantly crease nose} (top) and \textit{highly parted lips} (bottom), help human easily imagine of synthesized facial images. Thus, natural language is more intuitive for human to interpret the facial expression synthesis instead of directly conveying AUs. 
	
	Therefore, in this paper, we propose a novel facial expression synthesis model based on language description for expression. Our proposed model consists of two main categories:  Language-Visual Semantic Network (LVSN) and Facial Expression Synthesizing Network (FESN).
	
	The LVSN is designed to match language semantic features from language-based expression descriptions and visual semantic features from synthesized facial images. We effectively measure similarities between the language semantic features and visual semantic features with language-visual matching. Next, we devise the FESN to synthesize the input neutral facial image with the expression description by fusing the language semantic and the input face feature information. With those information, the output facial image is synthesized by multi-level synthesizers with expressive word attention module that can manipulate each muscle movement independently. Details of facial muscle movement is further guided by verification modules (\textit{e.g.,} discriminator). With these two networks being set, the proposed method can generate the facial image with detailed expressions, even minor expressions. To sum up, our main contributions are: 
	\begin{compactitem}
		\itemsep0em 
		\item We propose the facial expression synthesis from a single facial image using a guidance of language-based description which is humanly interpretable. To the best of our knowledge, it is the first time to directly use natural language when generating facial expression. 
		\item Through effective embedding the language semantic on the input facial features, it is possible to control detailed facial movements independently by manipulating descriptive words that represent the expression in detail.
	\end{compactitem}
	\section{Proposed method}
	Fig. 2 shows the overall architecture of the proposed method. It contains Language-Visual Semantic Network (LVSN) and Facial Expression Synthesizing Network (FESN). With language-based facial expression description, the LVSN and FESN  help synthesizing the input neutral image to well elaborate both major and minor expression. Detailed explanation is given in the following subsections. 
	
	\subsection{Language-Visual Semantic Network (LVSN)}
	
	As shown in the bottom part of Fig. 2, the LVSN is designed for guiding the language-based facial expression description and the synthesized facial image to be similar, using language-visual matching. This network can effectively embed the language and visual semantic features by matching the domain level of them. The LVSN consists of Language Semantic Encoder (LSE) and Visual Semantic Encoder (VSE). Based on the $N$-words of the facial expression description, the LSE outputs language semantic features $\mathbf{w}=\{w_{1}, w_{2}, ..., w_{N}\}\in \mathbb{R}^{512\times N}$ through bidirectional LSTM \cite{biLSTM}. The VSE receives synthesized facial image and extracts visual semantic features $\mathbf{f_{\mathrm{s}}}\in \mathbb{R}^{512}$. The objective function for language-visual matching is addressed in Section 2.3. 
	
	\subsection{Facial Expression Synthesizing Network (FESN)}
	
	The top part of Fig. 2 shows the network configuration of the FESN. It consists of three main modules: Facial Feature Encoder (FFE), Multi-level Expression Synthesizer (MES), and discriminators. The FFE module is devised to extract the input facial feature. When the neutral image $\mathbf{I_{\mathit{i}}}\in \mathbb{R}^{256 \times 256 \times 3}$ is given, the FFE extracts facial feature $\mathbf{f_{\mathit{f}}}\in \mathbb{R}^{2 \times 2 \times 512}$.
	
	The role of the MES is to generates a synthesized facial image $\mathbf{I_{\mathit{s}}}\in \mathbb{R}^{256 \times 256 \times 3}$. This module has three level expression synthesizers with residual blocks \cite{RESNET}, following the advantage of multi-level generator \cite{StackGAN}. The 1st level expression synthesizer receives concatenated feature $\left [ \mathbf{f_{\mathit{f}}};\hat{w}_{N} \right ]$, where  $\hat{w}_{N}\in \mathbb{R}^{2 \times 2 \times 512}$ is refined by the conditioning augmentation \cite{StackGAN} and resized to match the size of $\mathbf{f_{\mathit{f}}}$. Then, inspired by \cite{StackedGAN}, $n$-th expressive word attention module, $\textrm{att}_{n}$, is utilized. This module plays a role of focusing the local facial area with respect to every language semantic feature representing AU and its intensity. The expressive word attention module takes the language semantic feature $\mathbf{w}$ and the output facial feature of $n$-th level expression synthesizer $\mathbf{q_{\mathit{n}}}\in \mathbb{R}^{W \times H \times 512}$. Here, W and H are width and height of the facial feature, respectively. Each language semantic feature checks every sub-region of facial feature to find the region to attend with weight $\beta_{ij}$ as follows:
	\begin{equation}
	\alpha _{j}= \sum_{i=0}^{N-1}\beta _{ij}\hat{\mathbf{w}}_{i}, \:\: \textup{where} \:\:\: \beta _{ij}=\frac{\textrm{exp}\left ( \hat{\mathbf{q}}_{j}^{T} \hat{\mathbf{w}}_{i} \right )}{\sum_{k=0}^{N-1}\textrm{exp}\left ( \hat{\mathbf{q}}_{j}^{T} \hat{\mathbf{w}}_{k} \right )},
	\label{eq:1}
	\end{equation}
	where $i$ and $j$ are the index for word vector and sub-region of the facial feature, respectively. Here, $\hat{\mathbf{w}}$ is refined $\mathbf{w}$ followed by $1 \times 1$ convolution, and $\hat{\mathbf{q}}_{\mathit{n}}$ is the reshaped output facial feature ${\mathbf{q}}_{\mathit{n}}$ to match the shape of $\hat{\mathbf{w}}$. Then, the attention module $att$ produces the collection of all attended region, $\alpha _{0},\alpha _{1},...,\alpha _{N}$. With the expressive word attention module, the MES is able to control the detail facial movements independently by seeing attended word vector combination.
	
	Next, we design two discriminator modules: Verification Discriminator and Synthesis Discriminator. Motivated by multi-critic network \cite{Minho}, we devise the Verification Discriminator which consists of two discriminators: face verification discriminator and expression verification discriminator. The face verification discriminator $D_{FV}^{n}$ evaluates the expression synthesized image $\mathbf{I}_{s}^{n}$ and the target facial image $\mathbf{I}_{t}^{n}$. Moreover, the expression verification discriminator $D_{EV}^{n}$ distinguishes the difference of facial expression between the encoded $\mathbf{I}_{s}^{n}$ and the encoded $\mathbf{I}_{i}^{n}$ and the difference between the encoded $\mathbf{I}_{t}^{n}$ and the encoded $\mathbf{I}_{i}^{n}$, using convolutional encoder $f_{EV}^{n}$. $\mathbf{I}_{i}^{n}$ and $\mathbf{I}_{t}^{n}$ in each level are resized to match the sizes of the expression synthesized images, $\mathbf{I}_{s}^{1}\in \mathbb{R}^{64 \times 64 \times 3}$ and $\mathbf{I}_{s}^{2}\in \mathbb{R}^{128 \times 128 \times 3}$. This expression verification discriminator focuses on the expression variation by taking care of the details of facial muscle movement more carefully. Lastly, the Synthesis Discriminator is the same as the face verification discriminator, while the input is the final synthesized image. 
	
	\subsection{Objective Function for Network Training}
	
	Firstly, the language-visual matching loss $L_{LVM}$ is designed to make the language semantic features $\mathbf{w}$ and the visual semantic features $\mathbf{f}_{s}$ similar. We utilize multimodal similarity function \cite{Attngan} based on using cosine-similarity for $L_{LVM}$.
	
	Next, the loss for the face verification discriminator $D_{FV}$ for the synthesized facial images in $n=1,2$ can be written as:
	\begin{equation}
	\scalebox{0.91}[1]{$
		\begin{split}
		L_{FV}^{n} & =  -\frac{1}{2} \{  \textrm{log}  ( D_{FV}^{n}  (I_{t}^{n}) )  
		+ \textrm{log} ( 1- D_{FV}^{n} ( I_{s}^{n}) ) ) \} \\
		&-\frac{1}{2} \{ \textrm{log} ( D_{FV}^{n}( I_{t}^{n} ; \hat{w}_{N}  ))  
		+ \textrm{log} ( 1-D_{FV}^{n} ( I_{s}^{n} ) ; \hat{w}_{N} )  ) \},
		\end{split}$}
	\label{eq:2}
	\end{equation}
	where $I_{s}=G^{n} ( I_{i}^{n}; \hat{w}_{N})$. Here, $G^{n}$ is the $n$-th level generator. The first two terms represent an unconditional discriminator loss without language condition, and the last two terms show a conditional discriminator loss that conditioned on $\hat{w}_{N}$.
	
	The loss function for the expression verification discriminator $D_{EV}$ is shown as follows:
	\begin{equation}
	\scalebox{0.845}[1]{$
		\begin{split}
		L_{EV}^{n} &=  -\frac{1}{2} \{ \textrm{log} ( D_{EV}^{n}  ( F_{EV_{t}}^{n} )  
		+ \textrm{log}  ( 1- D_{EV}^{n}  ( F_{EV_{s}}^{n} ) )   \} \\
		&-\frac{1}{2} \{  \textrm{log}  ( D_{EV}^{n} ( F_{EV_{t}}^{n}; \hat{w}_{N}  )  )  
		+ \textrm{log}  ( 1-D_{EV}^{n} ( F_{EV_{s}}^{n} ; \hat{w}_{N} )  )  \},
		\label{eq:3}
		\end{split}$}
	\end{equation}
	where $F_{EV_{t}}^{n}=f_{EV}^{n}( I_{t}  )-f_{EV}^{n}( I_{i}  ) $ and $F_{EV_{s}}^{n}=f_{EV}^{n}( I_{s}  )-f_{EV}^{n}( I_{i}  )  $. Lastly, the synthesized discriminator loss $L_{syn}$ is the same as $L_{FV}^{n}$ shown in Eq. 2 with $n=3$.  Finally, the total discriminator loss is:
	\begin{equation}
	L_{D,total}=L_{syn}+\sum_{i=1}^{2}\left ( L_{FV}^{i}+L_{EV}^{i} \right ).
	\label{eq:4}
	\end{equation}
	\begin{figure*}[t!]
		\begin{minipage}[b]{1.0\linewidth}
			\centering
			\centerline{\includegraphics[width=14.5cm]{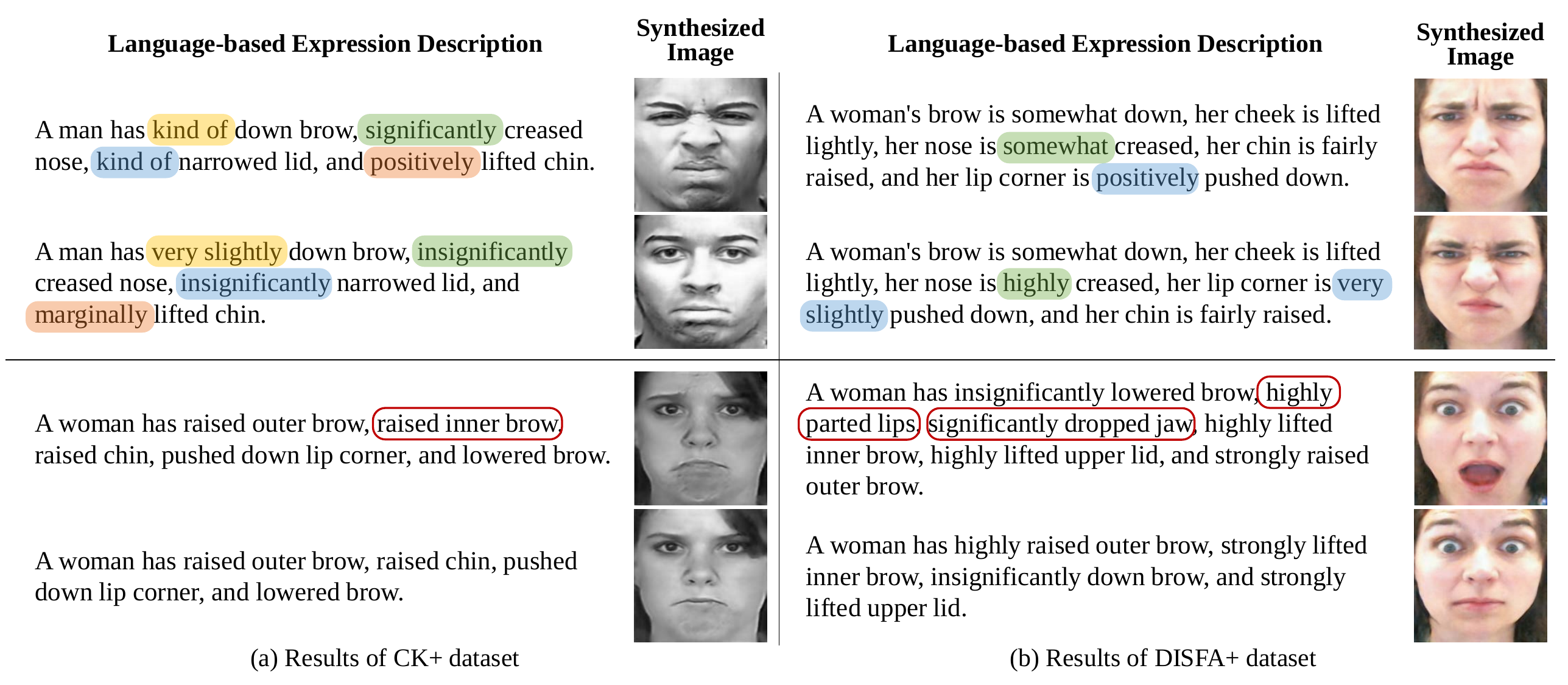}}
		\end{minipage}
		\vspace{-0.8cm}
		\caption{The synthesized facial expression image showing the manipulation of expressive word choice. The figure represents (a) results of CK+ dataset and (b) results of DISFA+ dataset. The top section controls the adverbs. The highlighted words are the adverbs to control. The bottom manipulates red circled word phrases by removing some descriptive words.}
		\label{fig:3}
		\vspace{-0.4cm}
	\end{figure*}
	
	For generator loss, we apply adversarial loss, identity loss, and reconstruction loss. The objective function for adversarial loss is defined as:
	\begin{equation}
	\scalebox{0.97}[1]{$
		\begin{split}
		L_{adv}^{m} =  -\frac{1}{2} \{   \textrm{log}  ( D_{FV}^{n}  ( I_{s}^{m}  )  )  
		+  \textrm{log}  ( D_{FV}^{n}  ( I_{s}^{m} ; \hat{w}_{N} ) ) \},
		\label{eq:5}
		\end{split}$}
	\end{equation}
	for $m=1,2,3$. Identity loss $L_{id}^{m}$ calculates L2 loss of output features of $\mathbf{I}_{t}^{m}$ and  $\mathbf{I}_{s}^{m}$ from the FFE, respectively, in each $m$-th level. Lastly, we use L1 distance between $\mathbf{I}_{t}^{m}$ and  $\mathbf{I}_{s}^{m}$ for reconstruction loss $L_{recon}^{m}$ in each $m$-th level. Therefore, the total generator loss is:
	\begin{equation}
	L_{G,total}=\sum_{i=1}^{3}\left ( \lambda _{1}L_{adv}^{i}+\lambda _{2}L_{id}^{i}+\lambda _{3}L_{recon}^{i} \right ),
	\label{eq:8}
	\end{equation}
	where $\lambda _{1}$, $\lambda _{2}$, and $\lambda _{3}$ are the hyper-parameters controlling relative importance of each loss. As a result, we sum all $L_{LVM}$, $L_{D,total}$, and $L_{G,total}$ for total loss function.
	
	\section{Experimental results}
	
	\subsection{Language-based Facial Expression Datasets}
	In experiments, we utilized CK+ \cite{CK+} and DISFA+ \cite{DISFA+} from the language-based facial expression description datasets \cite{FTDE} and revised them for diverse representations. The dataset was built to express a facial image with comprehensive aspects: gender, facial AUs, and following intensities. However, \cite{FTDE} only brought one sentence generation rule, so we created two more protocols for generating language-based sentence.
	
	CK+ contains $123$ subjects. There were $593$ peak frames containing highest intensities with both emotions and facial action units. If intensities are not listed, qualifying words (i.e., adverb) were not added to the sentence. We utilized $12$ subjects for evaluation. DISFA+ consists of $9$ subjects, where we trained on 8 subjects and evaluated on 1 subject. The total of $1,940$ images contain both facial AUs and intensities. Note that the identities used in the training phase and the testing phase were totally separated.
	
	\subsection{Experimental Setup}
	We pre-trained the LSE and the VSE with learning rate $1 \times 10^{-5}$ and fixed their weights. We use the Inception-v3 \cite{InceptionV3} for the VSE. Then, the weights of both encoders are fixed. We used Adam optimizer \cite{Adam} for all trainable models with learning rate $2 \times 10^{-4}$. The hyper-parameters of Eq. 6 were set to be $\lambda _{1}=1$, $\lambda _{2}=5$, and $\lambda _{3}=0.005$.
	
	\subsection{Qualitative Results}
	Fig. 1 briefly introduces the synthesized facial expression images as exemplars. The figure shows that the proposed method clearly follows the language-based description with identity preserving. More comprehensive results are demonstrated in Fig. 5 in supplementary materials.
	
	In addition, in order to verify that the model could manipulate the language-based expression description, we made modifications on words that describe the expressions. Fig. 3 indicates the manipulation of expressive word choice. In the top section of Fig. 3, the highlighted words are the adverbs to control, where the adverbs represent the intensities of facial expression. For instance, for the top left images, when we lessened the intensity of creased nose, \textit{significantly} to \textit{insignificantly}, the nose part of the synthesized image became less creased. This means that the proposed method could control the intensities of the facial muscle movement. 
	
	The bottom manipulates the facial expressions by adding or removing some descriptive words. For example, considering bottom right images pair, when we removed \textit{highly parted lips} and \textit{significantly dropped jaw} (red circled on the top image), the bottom image was synthesized without indicating \textit{parted lips} and \textit{dropped jaw}. As a result, these synthesized facial images show how well the network manipulates each word independently in synthesizing the image.
	
	\begin{table}[t!]
		\newcolumntype{C}[1]{>{\centering\arraybackslash}m{#1}}
		\label{table:1}
		\vspace{-0.3cm}
		\renewcommand{\tabcolsep}{2mm}
		\renewcommand{\arraystretch}{1.14}
		\caption{Quantitative Results on SSIM (higher is better) and Fr$\acute{\textrm{e}}$chet Inception Distance (lower is better).}
		\vspace{-0.05cm}
		\centering
		\begin{center}
			\resizebox{0.99\linewidth}{!}
			{
				\begin{tabular}{c|cc|cc}
					\hline \hline
					& \multicolumn{2}{>{\centering\arraybackslash}m{1in}|}{$\textbf{CK+}$} & \multicolumn{2}{>{\centering\arraybackslash}m{1in}}{$\textbf{DISFA+}$} \\ \cline{2-5} 
					& $\textbf{SSIM}$       & $\textbf{FID}$         & $\textbf{SSIM}$        & $\textbf{FID}$          \\ \hline
					\begin{tabular}[c]{@{}c@{}}Proposed Method\\ $\textit{w/out}$ language-visual matching\end{tabular}  & 0.201      & 149.148     & 0.237       & 78.252       \\ \hline
					\begin{tabular}[c]{@{}c@{}}Proposed Method\\  $\textit{w/out}$ expressive word attention\end{tabular} & 0.304      & 85.782      & 0.260       & 80.553       \\ \hline
					Proposed Method                                                                           & $\mathbf{0.661}$      & $\mathbf{57.950}$      & $\mathbf{\textbf{0.648}}$       & $\mathbf{\textbf{53.791}}$       \\  \hline\hline
					
				\end{tabular}
			}
		\end{center}
		\vspace{-8.0mm}
	\end{table}

	\subsection{Quantitative Results}
	Since our proposed method is the first study of synthesizing facial expression using language-based dataset, we performed SSIM \cite{SSIM} and FID \cite{FID} to verify whether the synthesized facial images are well generated compared to the target facial images. We compared our results with the output of the model without use of language-visual matching and without use of the expressive word attention module. Clearly, Table 1 shows that our overall architecture achieved high performance on generating the facial expression images with guidance of language-visual matching and expressive word attention.
	
	\section{Conclusion}
	
	In this paper, we introduced the novel facial expression synthesis that uses language-based facial expression description. Natural language helps humans understand and visualize conceptual context, so the guidance of language in synthesizing the facial image provides clear interpretability for humans. Through language and facial features matching, our proposed method can control detailed facial movements independently by manipulating descriptive words that represent the detailed expression. To the best of our knowledge, it is the first time to directly use language when generating facial expression.

		\begin{spacing}{0.95}
			\bibliographystyle{IEEEbib}
			\bibliography{refs}
		\end{spacing}

\end{document}